\documentclass[letterpaper]{article} 
\usepackage[submission]{aaai23}  
\usepackage{amsmath}
\usepackage{times}  
\usepackage{helvet}  
\usepackage{courier}  
\usepackage[hyphens]{url}  
\usepackage{graphicx} 
\usepackage{natbib}  
\usepackage{caption} 
\usepackage{lineno}
\usepackage{listings} 
\usepackage{xcolor} 

\definecolor{codegreen}{rgb}{0,0.6,0}
\definecolor{codegray}{rgb}{0.5,0.5,0.5}
\definecolor{codepurple}{rgb}{0.58,0,0.82}
\definecolor{backcolour}{rgb}{0.95,0.95,0.92}

\lstdefinestyle{mystyle}{
    backgroundcolor=\color{backcolour},   
    commentstyle=\color{codegreen},
    keywordstyle=\color{magenta},
    numberstyle=\tiny\color{codegray},
    stringstyle=\color{codepurple},
    basicstyle=\ttfamily\footnotesize,
    breakatwhitespace=false,         
    breaklines=true,                 
    captionpos=b,                    
    keepspaces=true,                 
    numbers=left,                    
    numbersep=5pt,                  
    showspaces=false,                
    showstringspaces=false,
    showtabs=false,                  
    tabsize=2
}

\usepackage{algorithm2e}
\usepackage{tikz}
\usetikzlibrary{calc}
\usepackage{booktabs}
\usepackage{multirow}
\usepackage{algorithmic}
\usepackage{multicol}

\title{Transformer Based Planning in the Observation Space with Applications to Trick Taking Card Games}
\author{%
  Douglas Rebstock$^1$, \textbf{Christopher Solinas}$^1$, \textbf{Nathan R. Sturtevant}$^{1,2}$, \textbf{Michael Buro}$^1$\\
  $^1$\textrm{Department of Computing Science, University of Alberta}\\
  $^2$\textrm{Alberta Machine Intelligence Institute (Amii)}\\
  \texttt{\{drebstoc,solinas,nathanst,mburo\}@ualberta.ca} \\
}

\begin{document}
\thispagestyle{plain} 

\maketitle

\begin{abstract}
Traditional search algorithms have issues when applied to games of imperfect information where the number of possible underlying states and trajectories are very large. This challenge is particularly evident in trick-taking card games. While state sampling techniques such as Perfect Information Monte Carlo (PIMC) search has shown success in these contexts, they still have major limitations.

We present Generative Observation Monte Carlo Tree Search (GO-MCTS), which utilizes MCTS on observation sequences generated by a game specific model. This method performs the search within the observation space and advances the search using a model that depends solely on the agent's observations. Additionally, we demonstrate that transformers are well-suited as the generative model in this context, and we demonstrate a process for iteratively training the transformer via population-based self-play.

The efficacy of GO-MCTS is demonstrated in various games of imperfect information, such as Hearts, Skat, and "The Crew: The Quest for Planet Nine," with promising results.
\end{abstract}
\section{Introduction}

In games of imperfect information, applying traditional search algorithms is often not feasible since the underlying state is unknown and the space of possible histories is too large. Trick-taking card games are examples of imperfect information games in which these issues often exist. Perfect Information Monte Carlo search (PIMC) \cite{levy1989million} has been shown to be effective in trick-taking card games \cite{long2010understanding}, however it is fundamentally limited as it solves sampled underlying states and not the original imperfect information problem.

We introduce Generative Monte Carlo Tree Search (GO-MCTS), a method that performs MCTS \cite{kocsis2006bandit} search in the observation spaces, with state transitions provided by a generative model. In this paper, we chose to use a relatively small transformer for the generative model. By doing search on this observation space and advancing the search using a model conditioned solely on the search agent's observations, we sidestep the need for knowing the underlying state. This change avoids common issues found in PIMC such as strategy fusion and non-locality \cite{frank1998search}, and greatly reduces the search space. Strategy fusion refers to how PIMC can use different strategies in different sampled worlds (not possible) and non-locality refers to how sub-tree values can be affected by values of nodes outside of the sub-tree, which is not the case for perfect information games. The approach is applicable to all games of imperfect information, however it is best suited to games where the observations are easily tokenized, whereas games with complex trajectories of raw inputs (such as video-games) would present difficulties. Transformers can be trained using only the raw observations as inputs, allowing the approach to be easily applied to new domains.

We also demonstrate how to train such a transformer, either bootstrapping from a uniform random player or from an existing weak policy. We do this through iterative policy improvement using Neural Fictitious Self-Play (NSFP) \cite{heinrich2016deep}, and parameterizing the population policies with the transformer. We provide experimental results for Hearts, Skat, and \emph{The Crew: The Quest for Planet Nine} \cite{sing2000crew}, all popular trick-taking card games. We show that in all of these games, GO-MCTS is able to directly improve upon the final trained policy from the iterative learning process, and provides new state of the art results in Hearts and the Crew.

\section{Problem Background} 

Within the broad domain of artificial intelligence (AI) research, games have traditionally been used as test beds for creating novel algorithms. One of the basic properties of a game is whether it features perfect or imperfect information . Games of perfect information have the property that all players have access to all the game information at all times. Well studied games such as Chess, Go, and Checkers all feature perfect information since the players can directly observe the current state of the board as well as all actions taken by the other players and actions are taken sequentially. 

\begin{figure}[b]
  \centering
  \tikzset{
    solid node/.style={circle,draw,inner sep=1.5,fill=black},
    hollow node/.style={circle,draw,inner sep=1.5}
  }
  \begin{tikzpicture}[scale=1.5,font=\footnotesize]
    \tikzstyle{level 1}=[level distance=10mm,sibling distance=30mm]
    \tikzstyle{level 2}=[level distance=10mm,sibling distance=10mm]
    \node(0)[solid node,label=above:{$P_1$}]{}
    child{node(1)[solid node]{}
    child{node[hollow node,label=below:{$(20,0)$}]{} edge from parent node[left]{$Left$}}
    child{node[hollow node,label=below:{$(0,2)$}]{} edge from parent node[right]{$Right$}}
    edge from parent node[left,xshift=-5]{$A$}
    }
    child{node(2)[solid node]{}
    child{node[hollow node,label=below:{$(1,0)$}]{} edge from parent node[left]{$Left$}}
    child{node[hollow node,label=below:{$(0,-500)$}]{} edge from parent node[right]{$Right$}}
    edge from parent node[right,xshift=5]{$B$}
    };
    \draw[dashed]($(1) + (-.2,.25)$)rectangle($(2) +(.2,-.25)$);
    \node at ($(1)!.5!(2)$) {$P_2$};
  \end{tikzpicture}
  \caption{An example of an imperfect information game. $P_2$ cannot distinguish between states inside the information set (dotted rectangle) since they could not observe the private move made by $P_1$.}
  \label{fig:imperfect}
\end{figure}
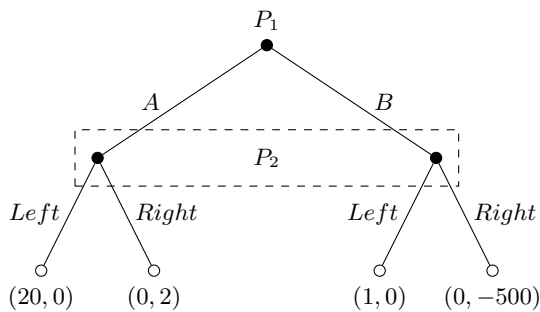

Tree search has been used to great effect in perfect information games. In tree search, a player evaluates diﬀerent sequences of moves that follow from the current state, and selects an action based on the result of this search. Tree search techniques have been used to solve Checkers \cite{schaeffer2007checkers}, and surpass human playing ability in Chess \cite{campbell2002deep} and Go 
\cite{silver2016mastering}. Simple state-based tree search cannot be naively applied in games of imperfect information, since the player may not always know what state they are in. Figure \ref{fig:imperfect} shows a simple example of an imperfect information game in which Player 2 ($P_2$) does not know whether they are in the left or right branch of the game tree since they do not observe the first action of $P_1$. Since they don't know what state they are in, they don't know which next state they will transition into for a given action, thus making the required tree traversal impossible. The collection of all states that the player thinks they may be in is termed the information set, and is indicated by the dashed box in the figure. This causes $P_2$ to have to consider all the branches in the tree and reason upon the strategy of $P_1$. While this often can be done in small games, this reasoning over possible states can be intractable when the information state is extremely large, which is often the case in the domain of trick-taking card games. While there are inherent difficulties, search has been used for games of imperfect information to great success in poker with DeepStack \cite{moravvcik2017deepstack} and Libratus \cite{brown2018superhuman}. The specific methods and algorithms that employ search in imperfect information games are quite varied and depend heavily on the intricacies of the game itself.

\subsection{Trick Taking Card Games}

Trick-taking card games are a popular category of card games that generally share certain features. The game is played as a sequence of hands in which each player is dealt a set of private cards. The hand itself is made up of the play of a number of tricks. The trick is the basic unit of play, as the players take turns placing one of their cards face up on the table. In most games once each player has played a card, the player with the highest valued card wins (takes) the trick. The outcome of the hand is tied to who won which tricks, however the scoring differs from game to game.

Since trick-taking card games feature private information, they are examples of imperfect information games. Due to this, and the large number of possible initial deals, the information set sizes are extremely large. For example, in the game of Hearts, 52 cards are dealt to four players, so there would be over \(5 \times 10^{26}\) deals and the size of the information set for a player after the initial deal would be over \(8 \times 10^{16}\). Each state would then have a set of possible trajectories that is exponential in the number of cards since all cards are sequentially played. In the worst case where all other players cannot follow the lead suit throughout the game, this would be \(12! \times 13!^3\), over \(1 \times 10^{38}\). Due to this, the size of the search space is too large for the direct application of tree search algorithms. The multiplayer aspect (usually more than two players) also provides unique challenges \cite{khan2002non}.  Also, the games can feature partial or full cooperation, which can provide different challenges than the more typical fully adversarial settings seen in most other well studied games.
\section{Generative Observation MCTS}

\begin{algorithm}[t]
\DontPrintSemicolon
\SetKwFunction{FMain}{GO-MCTS}
\SetKwFunction{FSelect}{Select}
\SetKwFunction{FSample}{Sample} 
\SetKwFunction{FBack}{Backup}
\SetKwFunction{FArgMax}{Argmax}
\SetKwFunction{FExpand}{Expand}
\SetKwFunction{FRollout}{Rollout}
\SetKwProg{Fn}{Function}{}{}

\Fn{\FMain{$h_{init}$, $Player$, $n$}}{
    $Tree \leftarrow \emptyset$ \;
    $root \leftarrow h_{init}$\;
    \For{$i \leftarrow 1$ \KwTo $n$}{
        $h \leftarrow root$\;
        \While{$h$ not Terminal}{
            \uIf{Player is to move}{
                \uIf{$h$ is in $Tree$}{
                    $a \leftarrow \FSelect{h, Tree}$ \;
                    $h \leftarrow h \;||\; a$ \; 
                }
                \Else{
                    $\FExpand{h, Tree}$ \;
                    $h \leftarrow \FRollout{h}$ \; 
                    break \;
                }
            }
            \Else{
                $o \leftarrow \FSample{h}$ \; 
                $h \leftarrow h \;||\; o$ \; 
            }
        }
        \FBack{$h, Tree$} \;
    }
    \KwRet $\FArgMax{Tree, root}$ \;
}
\caption{Generative Observation Monte Carlo Tree Search (GO-MCTS) algorithm. $h$ represents the observation history from the view of the current player, $a$ is an action, and $Tree$ is a set that maintains the summary statistics for the tracked observation histories. $Player$ indicates search player, $h_{init}$ is the root observation history of the search player, and $n$ is the number of runs. $\FSample$ uses the generative model to produce the next observation, which is appended to create the next observation history. $\FArgMax$ returns the action with the highest value. $\FSelect$ chooses the action with the highest score which is likewise appended to the history. A more detailed version is depicted in the supplemental material}
\label{alg:go-mcts}

\end{algorithm}

We present Generative Observation MCTS (GO-MCTS) in Algorithm \ref{alg:go-mcts}. In this proposed algorithm, we sidestep the issue of not knowing the true underlying state of the game by performing search in the observation space. GO-MCTS works by using an approximation of the observation dynamics model to perform MCTS in this generated observation space. We also provide the following theoretical background to provide intuition on the search.

When we formulate the problem in this way, we roll all other players and the environment into the dynamics model. Given fixed policies for the other players and fixed transition probabilities, there trivially exists fixed transition probabilities between underlying states. The problem can be formulated as a Partially Observable Markov Decision Process (POMDP) \cite{aastrom1965optimal}, since the transition only depends on the current state and the player cannot fully observe the states. By defining the state to be the observation sequence at the point in which the agent chooses an action, the problem can be represented as a regular MDP. This work follows from \cite{kaelbling1998planning}. In our analysis, we assume that the strategies to feature perfect recall and the games are of finite length.

Since the observation sequence is directly constructed by appending new observations, sequential queries of a generative model can be used to transition to the next state after the agent chooses an action. This method of using a generative model to implicitly represent an MDP is shown in \cite{kearns2002sparse} along with its usage in various solution concept. If the generative model produces a distribution of sequences that matches the true state distribution of the original MDP describing the game in question, solving for the implicit MDP will be sufficient.

Monte Carlo Tree Search (MCTS) has been shown to converge to find the highest value action for finite-horizon MDPs \cite{kocsis2006bandit}, so GO-MCTS will converge to this optimal solution in the limit if the generative model aligns with the true distribution. This connection was shown in \cite{silver2010monte}, however sampling discrete underlying states was used in that work. In practice though, the model will usually represent an approximation of the true observation dynamics. This can lead to compounding errors and arbitrarily bad results. Also, the cost to query such models can be very expensive computationally. Furthermore, the model corresponds to fixed policies, thus GO-MCTS is useful for best response against fixed opponents, not for directly solving for solution equilibria amongst strategic actors. And finally, we must have access to a reasonably good generative model, which is often not the case. 

Within the algorithm, any generative model would work as long as it could generate the next observation given the current observation state. In our application, we chose to use transformers due to their strength in sequence generation in Natural Language Processing (NLP). We reason that the game-play in trick-taking card games bears resemblance to natural language. For example, a card can be played to communicate information to another player, however the meaning needs to be derived in context with conventions and previously played actions. In the analogy of a game and sentence, the actions would be the words and the grammar would be the rules and conventions.

In the experimental implementation, a transformer is used for generating the next observation state, value initialization for expanded nodes, and value approximation for early stopping. While the transformer was used for the state transitions, the game simulator was used to determine the player to move, the final result, and move legality for the search agent. Since the generative model predicts observations of opponent actions without explicit knowledge of legal observations, the generative model is capable of generating impossible histories. For the terminal histories, we can retrospectively construct the original state and verify legality of the actions. If any actions were illegal or no original state could be constructed, we deem the history to be illegal and bias the search away from the nodes that were part of the run. This is important since the value estimation on illegal paths is very noisy, so the search can optimize towards these illegal trajectories. This procedure is part of \FBack and the full details can be found in the supplemental materials. Another way of avoiding illegal trajectories altogether is to only allow the generation of legal observations, however, generating the set of legal actions at each point has been shown to be generally intractable \cite{solinas2023history}, which is related to why we pursued the use of the generative model in the first place. 
\section{Related Work}

In this section, we highlight developments in search methodologies within imperfect information settings, particularly in trick-taking card games, and explored the integration of generative models in game planning and decision-making.

\subsection{Search in Trick Taking Card Games}

Determinized search is often used in the application of search based techniques in imperfect information settings. This approach samples a possible state from the player's information set, and then performs a search rooted at that state. PIMC is a form of determinized search in which all possible actions are evaluated by sampling states from the current information state, and averaging the values over the states which were determined using perfect information algorithms \cite{levy1989million}.  Algorithm \ref{alg:pimc} is a basic version of PIMC. 

\begin{algorithm}[t]
  \SetKwFunction{FMain}{PIMC}
  \SetKwFunction{sample}{Sample}{}
  \SetKwFunction{moves}{A}{}
  \SetKwFunction{weight}{ProbabilityEstimate}{}
  \SetKwFunction{norm}{Normalize}{}
  \SetKwFunction{perf}{PerfectInfoVal}{}
  \SetKwProg{Fn}{}{}{}
  \Fn{\FMain{\textrm{InfoSet} $I$, \textrm{int} $n$}}
  {
    \For{$a \in \moves{$I$}$} {
      $v[a] = 0$
    }
    \For{$i \in \{1..n\}$}
    {
      $s \leftarrow \sample{$I,p$}$ \;
      \For{$a \in \moves{$I$}$}
      {
        $v[a] \leftarrow v[a] + \perf{$s, a$}$
      }
    }
    \Return $argmax_a{v[a]}$
  }
  \vspace{0.5cm}
\caption{ Basic PIMC algorithm; $p$ is the probability of being sampled, $a$ is action, $v$ is value, and $n$ is the number of worlds evaluated.}
\label{alg:pimc}
\end{algorithm}

The first notable application of PIMC in a trick-taking card game was in Contract Bridge \cite{ginsberg2001gib}. This also represents the first major success in using determinized search in this domain. PIMC has shown great success in other trick-taking card games including Skat with the Kermit bot \cite{buro2009improving}, and Hearts with the xinxin bot (webdocs.cs.ualberta.ca/~nathanst/hearts.html). Both of these players represent the current state-of-the-art in their respective domains, and are used as the baseline players for experimentation in this paper.

Information Set Monte Carlo Search \cite{cowling2012information} is another determinized search algorithm, with applications in card games. It has many variants, however the basic premise is similar to PIMC in the sense that it relies on sampling underlying states from information sets. Counter Factual Regret (CFR) algorithms have shown to be very effective in certain card games such as poker \cite{moravvcik2017deepstack}, but have not found success in the trick taking domain. This is primarily due to the large size of the underlying tree in trick taking card games which in practice does not scale well with CFR.

One common thread in these approaches is the need to sample the underlying state. Due to the extremely large size of the information sets, sampling relevant states can be very costly, as the majority of all possible states can often be disregarded in the context of any sensible playing strategy. Furthermore, the challenge of even constructing or generating sample states has shown to be generally intractable \cite{solinas2023history}. This problem of sampling underlying states is general in nature and is present in many different solution methods across many domains.  While determinized search is the main method of applying traditional search methods in imperfect information games, another option is to directly search over an implicit representation of the game.

\subsection{Generative Search in Games}

The concepts introduced in this paper, particularly concerning Generative Observation Monte Carlo Tree Search (GO-MCTS), resonate with various notable works in the field of planning and decision-making in stochastic or hidden environments.

Dyna-Q \cite{sutton1990integrated} is a reinforcement learning (RL) algorithm that integrates planning, acting, and learning. The main idea behind Dyna-Q is to use experiences to update a model of the environment, and then use this model to simulate and learn from hypothetical experiences in addition to real interactions with the environment. This formulation has been popular and has been influential in learning and planning systems, but in this context the planning is used as a means of learning and not for online search. Incorporation of generative models into the planning processes, particularly in the domain of robotics, has been extensively surveyed and recognized for its potential \cite{deisenroth2013survey}.

Generative Adversarial Tree Search (GATS) \cite{azizzadenesheli2018surprising} uses a generative adversarial network (GAN) as a dynamics model to enable MCTS search. The algorithm was tested on select Atari games but was found to perform poorly.

MuZero \cite{schrittwieser2020mastering} uses a custom recurrent model to allow for MCTS search in the latent space. The model uses the past observed states as input to get the root hidden state and recurrently creates subsequent hidden states by using the selected action as the input. The hidden state is used as an input to the learned policy, value, and reward functions which are used in the adversarial search. A key point here is that the hidden state is not directly optimized for reconstruction loss but is learned in the process of optimizing for the policy, value, and reward losses. The way the algorithm was presented, it was not suitable for imperfect information games. Also, due to the deterministic nature outlined for producing the hidden states, the process was only suited for deterministic games.

Vector Quantised-Variational AutoEncoders (VQ-VAE) have been used to enable MCTS search in stochastic games \cite{ozair2021vector}. This approach involves taking past observations to create a discrete latent state.  A learned transition model then generates the next latent state given an action. This approach relies on separately training the auto-encoder to minimize reconstruction loss of the actual state, and the transition model between latent states. Furthermore, the stochastic games investigated either had a straightforward simulator-based approach in the case of chess or else had the complexity of the domain dominated by the problem associated with planning in video games. Similarly, stochastic MuZero \cite{antonoglou2021planning} uses a VQ-VAE with a codebook of a fixed size, such that each entry relates to a single one-hot vector. This allows them to do Gumbel sampling and simplifies the MCTS search. They also stress the idea of after-states, which are positions in the search between a decision point. The experiments were in 2048, Backgammon, and 9x9 Go, all games that do not feature hidden states.

Besides their effectiveness in NLP, transformers have demonstrated substantial potential in modeling arbitrary sequences. Works such as Decision Transformers \cite{chen2021decision} and Sequence Transformers \cite{janner2021offline} have shown how these models can be adapted for reinforcement learning. Both these models model the sequence of state, action, rewards which differs from this work, as we chose to model only the trajectory of observations from a single player's perspective and the focus was on the RL control problem. Transformers were used in conjunction with MCTS for SameGame, a single-player puzzle game \cite{yaari2022mctransformer}. This work is similar to previous work done on SameGame \cite{seify2020single} and is closer to the original AlphaGo approach, as the transformer is used to guide the perfect information state simulator. Using transformers for imitation learning can be seen in Othello-GPT \cite{li2022emergent} but it does not explore search-based methods using it.

\section{Learning Generative Model}

The second half of this paper focuses on producing the transformer to be used in GO-MCTS, and is separate from the search algorithm. Training such a transformer requires a large amount of data in the form of observation sequences from previously played games. If a large body of representative game data exists, it can be used directly to train the transformer or if representative policies are known and computationally inexpensive to run, the data can be created as a first step. If neither are available, we can learn the policies as a first step. We took this latter approach in this paper, as it allows for this method to be more flexible and applicable to novel domains. To be clear, we did not employ GO-MCTS in this learning phase.

\subsection{Transformer Architecture}

We used a PyTorch \cite{pytiorch} implementation of the GPT2 architecture \cite{radford2019language} from the Huggingface Transformers library \cite{wolf2019huggingface} as the base for the network architecture. The encoded sequence of observations by the agent are directly fed into the network, which it uses to predicts the next observation. A secondary classification head was added to the model to predict the outcome score of the game. This output is analogous to the value head commonly seen in other policy-value predictor models. However, instead of predicting a real value, the model outputs the outcome probability distribution. This can be fed into the agents value function. We set a fixed value to each outcome, and thus the observation state value function is simply defined by: 
\begin{equation} \label{eq:Vpi(s)}
V(O) = \sum_{o}p(o|O)*v(o)
\end{equation}
where $v$ is the user defined outcome value function, $O$ is the observation sequence, $o$ is the outcome, $p$ is the probability of the outcome from the network. This was done to avoid issues we have encountered with regressors in non-Gaussian data as well to add flexibility in the value function for the agent. This flexibility can be used in reward shaping during the learning process or even during online search, however we did not explore these avenues here. 

Using Neural Networks to predict the next move or current value from a player's observations typically involves expert handcrafted encoding. In our case, we bypassed this by using the observation sequence directly as the input. However, we still had to define what constitutes an observation for the game. For example, being dealt a Queen of Diamonds is very different from seeing the player to the left of you playing the same card. In this case, we opted for encoding the card only, and left the resulting observation sequence to be understood through context. The next observation state is trivially encoded by appending a token corresponding to the new observation to the previous observation state encoding. At each iteration's training step, the most recent model is used to initialize the parameters. The training loss consists of the next observation prediction loss and the final outcome prediction loss. We use cross entropy loss for both and weighted the observation loss at 0.9 and the outcome loss at 0.1. The heavier weighting of the observation loss is done with the intuition that the prediction outcome is more of an auxiliary task in modelling the dynamics. We used the default AdamW optimizer \cite{loshchilov2017decoupled} Further details on architecture and training can be found in the supplemental materials.

\begin{table}[t]
  \caption{Number of observation encoding tokens for Hearts, Skat, \emph{The Crew}.}
  \label{hearts-encoder-table}
  \centering
  \begin{tabular}{clc}
    \toprule
    Game & Token Type & Number\\
    \cmidrule(lr){1-3}
    \multirow{4}{*}{Hearts} & Cards  & 52 \\
    & Positions & 4 \\
    & Pass Directions & 4 \\
    & \textbf{Total} & 60 \\
    \cmidrule(lr){1-3}
    \multirow{6}{*}{Skat} & Cards  & 32 \\
    & Bids / Replies & 68 \\
    & Pickup/Hand & 2 \\
    & Game Declarations & 7 \\
    & Declaration Modifiers & 4 \\
    & \textbf{Total} & 113 \\
    \cmidrule(lr){1-3}
    \multirow{9}{*}{\emph{Crew}} & Cards & 40 \\
    & Positions & 4 \\
    & Allocation & 4 \\
    & Tiles & 11 \\
    & Special & 18 \\
    & Communication & 4 \\
    & Replies & 2 \\
    & Radio Comms & 4 \\
    & \textbf{Total} & 83 \\
    \bottomrule
  \end{tabular}
\end{table}

\subsection{Learning through Self-Play} 

In order to produce the trained transformer, we employed iterative policy improvement using neural fictitious selfplay \cite{heinrich2016deep} using a population based approach \cite{bansal2017emergent}. The basic idea is to parameterize a policy using the transformer network, and to perform iterative policy improvement. The initial batch of data was produced by either a computationally inexpensive baseline search player or a uniform random player. We then trained the transformer network on this data-set. While the architecture was not the focus of the paper, trial and error was used to set parameters to jointly optimize for memory footprint, training time, and time to run an inference pass. 
For the subsequent iterations, we set one player to greedily select the action with the highest action-value estimate from the most recent network, and the rest of the players were set to sample from a policy previous iteration. This was done to avoid instability issues.

The greedy policy uses the trained network to select the highest valued action, so long as the predicted move probability is above a given threshold. For the remainder of the paper we will refer to it as the ArgmaxVal* policy, and its formulation is given as

\begin{equation}
\pi_{ArgmaxVal*}(h,a) = argmax(V_L(h,a) \\
\end{equation}
where
\vspace{-0.3cm}
\begin{equation}
  V_L(h,a;\lambda) = \left\{
     \begin{array}{@{}ll}
	V(h,a)  & \text{if} p_{legal}(a|h) \geq \lambda \\
       -\infty & \text{otherwise} \\
     \end{array}
   \right.
\end{equation} 

in which $p_{legal}$ is the probability normalized over all legal actions, $h$ is the state defined by the observation history, $a$ is the action, and $\lambda$ is the user defined threshold.

\section{Experimentation}
In this section, we detail the empirical investigations conducted to assess the effectiveness of Generative Monte Carlo Tree Search (GO-MCTS) in navigating the complex decision spaces of trick-taking card games, specifically Hearts, Skat, and \textit{The Crew: The Quest for Planet Nine}.

\subsection{Hearts}

Hearts is a 4 person trick taking card game in which players try to collect the least amount of points. It is commonly played such that the player with the lowest score when a single player hits 100 points is the winner. All cards of suit hearts are worth one point, and the queen of spades is worth 13 points. In this variant, the player with the two of clubs goes first and if a single player collects all the point cards (shooting the moon), they receive negative 26 points and all other players are awarded 0 points. We did not award points for Jack of diamonds or avoiding all tricks, both common variants. Full rules for hearts can be found at www.pagat.com/reverse/hearts.html. We defined an outcome by the allocation of points among the players, which resulted in 2234 unique outcomes.

This game is quite large, as 52 cards are dealt to 4 different players, and then all cards are sequentially played. Additional complexity comes from the initial phase of passing, in which all players will choose 3 cards to pass in a prescribed direction (alternates between right, left, across, and no pass).

\subsubsection{Training}

We built upon the Hearts implementation in the OpenSpiel repository \cite{LanctotEtAl2019OpenSpiel}. We create an agent in that could directly interface with the existing code and built a tournament manager to create game experience. 

We created 4,000,000 legal random play-outs and used the corresponding observation sequences to train the base network. We then performed 10 iterations of experience generation and retraining of the network. In the first experience generation, we set all players to follow the ArgMaxVal* policy. We did this to aggressively move the learning process into a space that resembled sensible play. The outcome value function, $v(o)$, for the player was set to
\begin{equation} \label{eq:v_hearts)}
v(o)_{hearts}=\frac{-{cp}_{player} + \overline{cp}_{opponents}}{26}
\end{equation}
where $cp_{player}$ is the card points of the search player and $\overline{cp}_{opponents}$ is the average card points of all other players. We chose to normalize it by a factor of 26, which is the total card points available to be allocated. 

If we were to assume the opponent points are spread evenly amongst themselves in expectation, then this value would represent the difference between the players score and their opponents in the limit. This is a simplification for our purposes since it matters which opponent gets the points taking into account termination when a player reaches 100 points.

In the subsequent generation steps, we set one player to use the ArgMaxVal* policy, and the remaining players were set to sample from the imitation policy parameterized from a previous iteration. The iteration for each imitation player was selected uniform randomly and was used for the entirety of the hand. We only used the observation data for the ArgMaxVal* players when training the next iteration. We did this to increase stability of the training process and avoid narrow exploitation of a single policy. Each generation step produced 500,000 hands. Further experimentation would be needed to determine the effects of each of these choices, but this was not the focus of this paper. Further details can be found in the supplemental material.

\subsubsection{Performance}

\begin{table}[t]
  \caption{Baseline results for hearts experimental players against the baseline xinxin bot}
  \label{hearts-results-table}
  \centering
  \begin{tabular}{lcccc}
    \toprule
    & All     & $3A/1B$ & $2A/2B$ & $1A/3B$ \\
    \midrule
    ArgmaxVal* (A) &  4.02 & 4.46 & 3.98 & 3.63 \\
    Xin-Xin (B) & 4.97 & 5.51 & 4.96 & 4.42  \\
    $\Delta$    & -0.95 & -1.05 & -0.98 & -0.79  \\
    \midrule
    GO-MCTS (A) & 3.89 & 4.53 & 3.80 & 3.39  \\
    Xin-Xin (B) & 5.64 & 6.52 & 5.61 & 4.79  \\
    $\Delta$    & -1.74 & -1.99 & -1.81 & -1.39  \\
    \bottomrule
  \end{tabular}
\end{table}

We chose the xinxin bot as the baseline player since it represents the current state-of-the-art in computer Hearts AI. Despite the absence of a dedicated publication detailing its construction, the bot's development is informed by key research by its author in several pertinent areas: challenges in multiplayer games \cite{sturtevant2004current}, opponent modeling \cite{sturtevant2006prob}, learning feature representation \cite{sturtevant2007feature}, and analysis of Upper Confidence Tree (UCT) search in multiplayer games \cite{sturtevant2008analysis}. It utilizes PIMC with evaluations done using UCT. It was made available within the OpenSpiel framework \cite{LanctotEtAl2019OpenSpiel}. We set its parameters to 2000 runs, 50 worlds, and C to 0.4, which is the default highest setting in its implementation. We chose to do a 14-way tournament in which every seating permutation of Player A (the experimental bot) and Player B (xinxin) are seated for a given deal, with the exception of all A or all B. Neither player was designed to change its play based on the identity of the other players. We ran a 3000 match tournament with the experimental bot using the ArgMaxVal* policy paramterized by the final iteration of the trained transformer. All results reported were found to be significant ($p<0.001$) using the Wilcoxon Ranked sign test. As indicated in Table \ref{hearts-results-table}, the ArgMaxVal* player outperforms the xinxin bot by 0.95 points on average, with further breakdowns provided in the table. When both are ran in the single thread settings we used for experimentation, the ArgMaxVal* player spends only 71 ms per turn whereas the xinxin bot spends much longer at around 2.9 seconds per turn. The 0.95 points advantage is extremely large in the context of Hearts, as it would equate to a 19.1 point difference when extrapolated in games to 100 points.

We repeated the same procedure, but now with the GO-MCTS player. Again, we used the final iteration of the transformer, and the same value function as the ArgMaxVal* player. From the Table \ref{hearts-results-table}, we can see the performance of the experimental player increased when we switched from the ArgMaxVal* player to the GO-MCTS player, performing 1.74 points better than the baseline, and 31.0 points when extrapolated to playing to 100 points. The increased performance came at a cost of greatly diminished speed, as it took around 25.6s per turn. From this work we believe that this GO-MCTS implementation represents the new state of the art in computer Hearts AI. Further details on the hyper parameters of the experimental bots can be found in the supplemental material.

\subsection{Skat} 

Skat is a three person trick taking card game, popular in Germany. It features a shortened 32 card deck, and three phases of gameplay: bidding, declaring, and cardplay. After each player is dealt 10 cards, players sequentially bid higher game values until two of the players have passed, i.e. they stopped bidding. The high bidder becomes the soloist and plays against the other two players. The soloist may pick up two hidden cards (the skat) and then secretly discards two cards to get down to 10 and then declares a game type of a greater or equal value to their bid. In the cardplay phase, the player tries to achieve the game type victory conditions through a series of tricks. The three main game types are suit, grand, and null. In suit and grand games the Jacks are trump cards, and the winning condition is usually to get the majority of card points. In addition to the Jacks, the suit the soloist chooses is also trump. The other main game type is null, in which there are no trump and the soloist wins if they can win no tricks. Full rules can be found at www.pagat.com/schafkopf/skat.html. Due to the size of the game, and the added complexities due to bidding, declaration, and discards, there are no strong full game search players. Kermit, the baseline player, uses PIMC only for the cardplay and uses a variety of other methods for the bidding, declaration, and discarding.

\subsubsection{Training}

Like Hearts, each possible allocation of tournament points over the three players were assigned as a unique outcome. Tournament points refers to the total points that a player would receive for the played hand under the standard tournament rules. For Skat, this works out to be 397 unique outcomes. The outcome value function was defined as

\begin{equation} \label{eq:v_skat)}
v(o)_{skat}=\frac{{TP}_{player}(o) - \overline{TP}_{opponents}(o)}{100}  
\end{equation}
where $o$ is the outcome, ${TP}_{player}$ is the tournament points awarded to the player and $\overline{TP}_{opponents}$ is the average tournament points awarded to the opponents. In this context, opponents refer to all other players. This was done with the same rationale as Hearts, with the dividing by 100 being an arbitrary choice to scale the values down. We encoded the observations in a similar manner to Hearts, but the added complexities in Skat necessitated a richer vocabulary as seen in Table \ref{hearts-encoder-table}.

We created the initial batch of data using a mixture of uniform random players and XSkat \cite{gerhardt_2004} bots, a weak scripted player. 4,000,000 games were produced using all permutations of the two bots. We originally attempted to use only uniform random players, however we found that the learning process was unstable. We believe that this was due to the high degree of heterogeneity, the complicated bidding process, and the reward structure. For example, a player can bid any of almost 70 values, with only a small fraction of these being suitable. And again, randomly selecting a sensible game type and discard combination is extremely improbable. To make matters worse, the cardplay dynamics hinge greatly on the aforementioned decisions. Due to the asymmetry between defender and soloist and players seated position, we found that the learning process quickly collapsed as the players learned to not bid at all, as the reward is simply 0 for all players. This negative result should remind the reader that the process in the current form can perform poorly and still requires adaptation to the individual game itself.

We used the same base model architecture as described for Hearts and followed the same iterative training approach, however, we continued to 20 iterations, but kept old iterations from the self-play pool up to a maximum of 10 iterations. We choose to do this since the quality of the player was quite poor after 10 iterations, and we hoped that dropping old policies would help the learning process. Again, producing the trained transformer was not the focus of this paper, so further experimentation would be necessary to validate these decisions. Further details can be found in the supplementary material.

\subsubsection{Performance}

\begin{table}[b]
  \caption{Baseline results for skat experimental players against the baseline Kermit bot}
  \label{skat-results-table}
  \centering
  \begin{tabular}{lccc}
    \toprule
    & All     & $1A/2B$ & $2A/1B$  \\
    \midrule
    ArgmaxVal* (A) &  14.12 & 16.45 & 9.44 \\
    Kermit (B) & 30.42 & 33.46 & 28.91  \\
    $\Delta$    & -16.31 & -17.01 & -19.46  \\
    \midrule
    GO-MCTS (A) & 17.97 & 19.35 & 15.20  \\
    Kermit (B) & 27.81 & 30.02 & 26.70  \\
    $\Delta$    & -9.84 & -10.67 & -11.50  \\
    \bottomrule
  \end{tabular}
\end{table}

We evaluated the players after the final iteration against the Kermit baseline player.  We chose to do a six-way tournament in which every seating permutation of players A and B are seated for a given deal, with the exception of all A or all B. We used standard tournament scoring and again, the identity of the players are not revealed. We ran a 3000 match tournament with the experimental bot using the ArgMaxVal* policy and the final iteration of the trained transformer, with the results indicated in Table \ref{skat-results-table}. The baseline Kermit player outperforms ArgMaxVal* player by 16.31 points on average. When both are ran in the single thread settings we used for experimentation, the ArgMaxVal* player spends 72ms per turn whereas the Kermit bot spends around 0.55 s per turn. All the results were found to be statistically significant, and furthermore, the 16.31 points advantage is extremely large in the context of Skat.

We repeated the same procedure, but now with the GO-MCTS player. Again, we used the final iteration of the transformer, and the same value function as the ArgMaxVal* player. As seen in the results from Table \ref{skat-results-table}, the GO-MCTS player performs 9.84 points worse than Kermit, however this is a 6.47 points improvement over the ArgMaxVal* player. The increased performance came at a cost of greatly diminished speed, as it took around 42 seconds per turn. Even with the large performance boost when applying GO-MCTS, the baseline player is clearly much stronger. All values reported were found to be significant ($p<0.001$) using the Wilcoxon Ranked sign test. Further details can be found in the supplementary material.

\subsection{\emph{The Crew: The Quest for Planet Nine}}

\emph{The Crew: The Quest for Planet Nine} is a popular cooperative trick taking card game published in 2019. While the game allows between 2-5 players, the default rules assume the 4 player case, therefore this is the version we used. The objective of the game is to win all 50 missions in the fewest number of attempts possible, however, each mission can be played in a standalone fashion. 

The 40 card deck features four colored suits (Pink, Blue, Green, Yellow) each with cards numbered 1 through 9, and a default highest ranking suit (Rockets) with cards numbered 1 through 4. To start a mission, the cards are dealt evenly between the 4 players, and the player that received the 4 of Rockets becomes the commander of the mission. The players attempt to take the tricks in such a way that achieves the mission objective. If they do so successfully, they all win. Otherwise they must repeatedly attempt the mission until they succeed.

Each mission is different, however many share common features. The most typical mission would start with the allocation of tasks followed directly by the game play. A task is completed when the assigned player wins the task's corresponding card in a trick. The method of task allocation, the number of tasks, the order in which tasks must be completed, and rules on communication can vary from mission to mission. While around 30 missions follow this general formulation, the remaining missions feature either unique modifications to this blueprint or diverge completely. Official game rules are provided on the publishers website at store.thamesandkosmos.com/products/the-crew.

\begin{figure*}[t]
  \centering
  \includegraphics[width=0.5\linewidth]{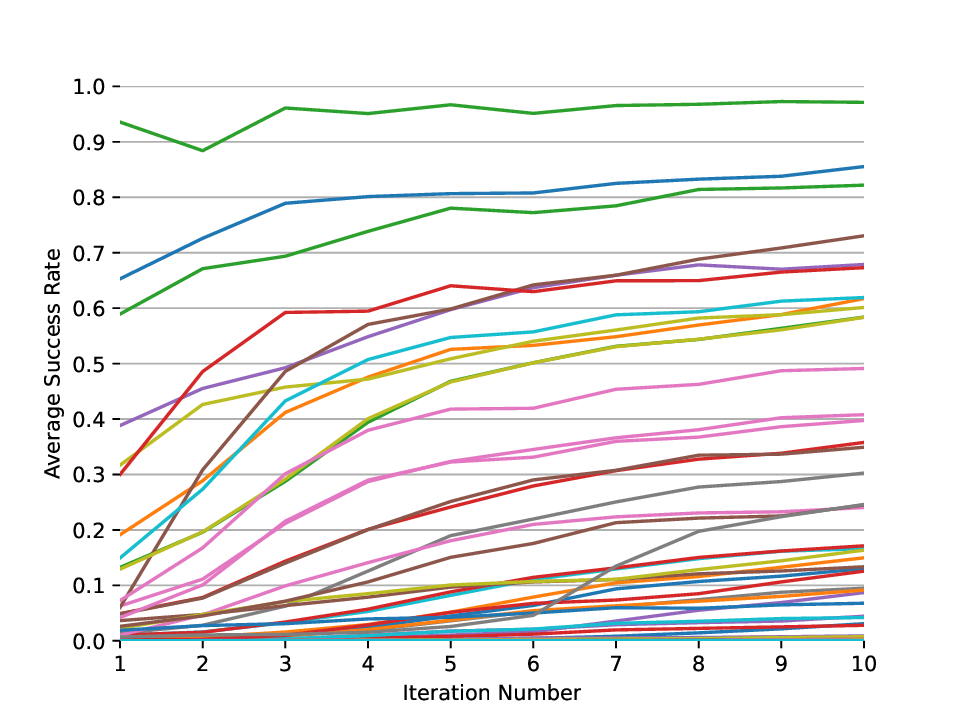}
  \caption{Average success rate for each mission in each training iteration (not using search).}
  \label{fig:crew_iterations}
\end{figure*}

\begin{figure*}[t]
  \centering
  \includegraphics[width=0.5\linewidth]{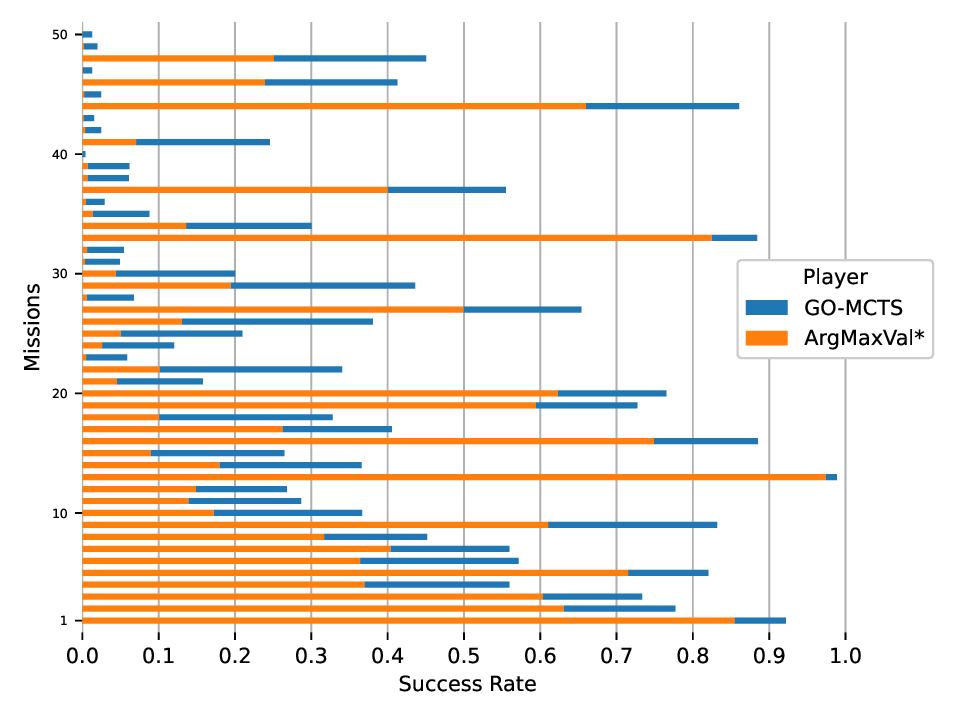}
  \caption{Average success for the final iteration of the ArgmaxVal* and GO-MCTS players. Results broken down for each mission.}
  \label{fig:crew_final}
\end{figure*}

\subsubsection{Training}

Due to the rules being different for each mission, in addition to the observation sequence, we included a unique prefix corresponding to the levels rules. Within the same formulation, one could interpret the prefix of tokens as series of chance moves by the world in the same manner as the deal. We decided to encode the mission using a custom language for the common features with the addition of a unique token for the remaining ones. The intuition behind this choice was to enable the learning agent to generalize knowledge between missions, as well as aid in the simplicity of the approach as it necessitates a single transformer. All relevant game information is encoded sequentially into the beginning of the sequence, including the mission encoding, the players position, the commanders position, as well as their dealt cards. Like the other games, the sequence corresponding to the next observation state is produced by appending the new observation token to the previous sequence. The choice of token vocabulary is indicated in Table \ref{hearts-encoder-table} .

Success or failure are the only two outcomes, so we assigned a value of 1 for all players when the team succeeded and 0 when they failed. We trained the initial network on a set of 10,000,000 games generated by a uniform random player. However, due to the sparsity of successful trials in the more difficult missions, we generated more games until we had a higher number of successful examples. This is an indication that this learning method would likely not scale with difficulty, as the amount of trials to randomly achieve even a single success could become prohibitively large.

One major difference in the training was that we chose to have all the agents using ArgmaxVal* policy from the most recent iteration. This choice was motivated by the fact that all players are working cooperatively, so we reasoned that the optimization process would be smoother than in the adversarial case seen in Hearts and Skat. Also, initial testing indicated initial learning improvements with this method. Our goal was to    simply train a transformer to test the GO-MCTS efficacy. For game play generation, we sampled all 50 missions with uniform probability. We performed 10 iterations, with 2,000,000 games per iteration. Figure \ref{fig:crew_final} indicates the relative success rate for each of the 50 missions over the 10 iterations. We set the threshold value to 0.05. We can see the general improvement in the success rate, and it is clear that there was still an improvement trend when the training ended.

\subsubsection{Performance}

Since this was a fully cooperative game without access to a baseline player for comparative analysis, we evaluated the performance based on the success rate, segmented by mission number. We conducted tests on both the ArgmaxVal* and the GO-MCTS players, executing each level 3,000 times to ensure statistical relevance. The outcomes, as depicted in Figure~\ref{fig:crew_final}, highlight the superior performance of the GO-MCTS player across all missions when compared to the ArgmaxVal* player. To ensure fairness in evaluation, pre-cardplay chance events were standardized for both players. A statistical analysis utilizing the Wilcoxon Signed-Rank test confirmed the significance of the observed differences, with \(p<0.001\). However, this enhanced performance of the GO-MCTS player comes with increased computational demands, averaging 5.9 seconds per turn, in stark contrast to the 41 milliseconds per turn required by the ArgmaxVal* player. For a comprehensive overview of the configurations used for both the search and greedy player, refer to the supplemental material.

\section{Discussion}

In all three domains, we clearly saw that we were able to improve the performance of the trained network player by performing local search through GO-MCTS. For Hearts, we surpassed the strength of the strong baseline player through an inexpensive one-step greedy policy ($ArgMaxVal*$) and provided the new state-of-the-art with our GO-MCTS player. In Skat, we were not able to surpass the strength of the baseline player, however we were able to produce a medium strength player and illuminate possible issues in the approach that would need to be further explored. And for \emph{The Crew}, we were able to create a strong baseline player in a complicated game and provide a basis for further research to build on. The strength of the GO-MCTS player is below that of an average experienced player, but we believe that surpassing this is doable within this framework without major changes. Over all domains, the GO-MCTS player used substantially more time per move than the $ArgMaxVal*$ or baseline player. However we believe the time per turn could be substantially reduced through simple but time consuming optimizations for the code. 

In Hearts and \emph{The Crew}, we were able to bootstrap the learning process off of random play-outs. While we could have bootstrapped existing policies for Hearts, starting from uniform random shows that the learning process is not limited to refining and exploiting existing policies. And for \emph{The Crew}, the choice was out of necessity as we did not have access to any expert data or desire to produce handcrafted player. As we move to more and more complex domains, this necessity will be the norm.

While the approach in this paper was shown to have merit, there is much to explore on how to make all aspects of the process more effective. The choice for the network architecture, observation encoding, learning algorithms and targets, batch sizes, and many other hyper-parameters makes a large difference in the efficacy of the training process. And for the GO-MCTS algorithm itself, the performance relies heavily on seemingly arbitrary parameter settings. Further refinement and exploration over these degrees of freedom could provide substantial improvements in both the performance and understanding of the process and how it relates to the domains in question.

\section{Conclusion}

In this paper we introduced a practical approach to searching over the observation space in imperfect information games through GO-MCTS. We provided our intuition on why a transformer is well suited in dealing with sequences in trick-taking card games, and demonstrated effective usage of a transformer as a generative model that could fit seamlessly into existing search paradigms. We also demonstrated an iterative neural fictitious population based self-play method to train the transformer, and provided experimental results in three large and complex trick-taking card games. While we got the process to work in Skat by bootstrapping off of a weak rule based policy, the resultant GO-MCTS player performed worse than the baseline. In Hearts and \emph{The Crew}, we provided new state-of-the-art results and were able to do so while bootstrapping off of uniform random play-outs.

\subsection{Future Work}

We believe further work in optimizing the implementation and further refinement of the process and hyper parameter choices could vastly improve the strength of the players in the trick-taking games studied while making them run much more efficiently. This would be a necessary step if we were to focus on the interplay between humans and bots in trick-taking card games, an area of possible future work.

We limited the domain to trick-taking card games, however this approach could be used in many imperfect information settings. A next step would apply this process past trick-taking games, in order to demonstrate generality and characterize the domains and conditions that it is effective. 
 
We would also like to investigate how to effectively incorporate GO-MCTS into the learning process and how to form the population dynamics and the learning updates to effectively utilize this additional computation while maintaining stability. Also, we would like to standardize the learning approach and perform ablations to determine the effects of our design choices.


\onecolumn
\section{Supplemental Materials}

\subsection{Model Architecture}

The custom GPT2PolicyValue model was built on-top of the GPT2PretrainedModel.

Modifications in the initialization:
\begin{lstlisting}[language=Python]
class GPT2PolicyValueModel(GPT2PreTrainedModel):
    def __init__(self, config, num_labels):
        super().__init__(config)
        self.num_labels = num_labels
        self.transformer = GPT2Model(config)
        self.lm_hidden = nn.Linear(config.n_embd, config.n_embd, bias=False)
        self.lm_head = nn.Linear(config.n_embd, config.vocab_size, bias=False)
        self.val_hidden = nn.Linear(config.n_embd, config.n_embd, bias=False)
        self.val_head = nn.Linear(config.n_embd, self.num_labels, bias=False)
        self.init_weights()
\end{lstlisting}

Modifications in the forward pass:
\begin{lstlisting}[language=Python]
        lm_logits = self.lm_hidden(hidden_states)
        lm_logits = F.gelu(lm_logits)
        lm_logits = self.lm_head(lm_logits)
        val_logits = self.val_hidden(hidden_states)
        val_logits = F.gelu(val_logits)
        val_logits = self.val_head(val_logits)
\end{lstlisting}

Modifications in the loss calculation:
\begin{lstlisting}[language=Python]
    loss_fct = CrossEntropyLoss()
    loss_lm = loss_fct(shift_logits.view(-1, shift_logits.size(-1)), shift_labels.view(-1))
    shift_val_logits = val_logits[..., :-1, :].contiguous()
    shift_val_labels = val_labels[..., 1:].contiguous()
    loss_val = loss_fct(shift_val_logits.view(-1, shift_val_logits.size(-1)), shift_val_labels.view(-1))
    loss = (loss_val  + loss_lm * 9) / 10
\end{lstlisting}

\begin{table}[h]
\centering
\caption{Detailed Model Configuration Parameters}

\label{tab:model_config}
\begin{tabular}{|l|l||l|l|}
\hline
\textbf{Parameter}                           & \textbf{Value}                & \textbf{Parameter}                           & \textbf{Value}                \\ \hline
activation\_function                         & gelu\_new                     & layer\_norm\_epsilon                         & 1e-05                         \\ \hline
attn\_pdrop                                  & 0                             & model\_type                                  & gpt2                          \\ \hline
bos\_token\_id                               & None                          & n\_embd                                      & 256                           \\ \hline
embd\_pdrop                                  & 0.05                          & n\_head                                      & 8                             \\ \hline
gradient\_checkpointing                      & false                         & n\_inner                                     & 1024                          \\ \hline
initializer\_range                           & 0.02                          & n\_layer                                     & 8                             \\ \hline
scale\_attn\_by\_inverse\_layer\_idx         & false                          & reorder\_and\_upcast\_attn                   & false                         \\ \hline
scale\_attn\_weights                         & true                          & resid\_pdrop                                 & 0.05                          \\ \hline
summary\_activation                          & null                          & summary\_proj\_to\_labels                    & true                          \\ \hline
summary\_first\_dropout                      & 0                             & summary\_type                                & cls\_index                    \\ \hline
summary\_use\_proj                           & true                          & torch\_dtype                                 & float32                       \\ \hline
transformers\_version                        & 4.12.5                        & use\_cache                                   & true                          \\ \hline
\end{tabular}
\end{table}

\begin{table}[h]
    \centering
    \begin{minipage}{.5\linewidth}
      \centering
        \caption{Game Specific Model Configuration Parameters for Hearts, Skat, and Crew}
        \label{tab:model_config_games}
        \begin{tabular}{|l|l|l|l|}
        \hline
        \textbf{Parameter} & \textbf{Hearts} & \textbf{Skat} & \textbf{Crew} \\ \hline
        vocab\_size        & 61              & 119           & 103           \\ \hline
        n\_positions       & 75              & 145           & 128           \\ \hline
        n\_ctx             & 75              & 145           & 128           \\ \hline
        eos\_token\_id     & 60              & 118           & 102           \\ \hline
        n\_labels          & 2237            & 397           & 2             \\ \hline
        \end{tabular}
    \end{minipage}%
    \begin{minipage}{.5\linewidth}
      \centering
        \caption{Additional Training Parameters}
        \label{tab:training_params}
        \begin{tabular}{|l|l|}
        \hline
        \textbf{Parameter}          & \textbf{Value}    \\ \hline
        Optimizer                   & AdamW             \\ \hline
        Learning rate               & \(1 \times 10^{-4}\) \\ \hline
        LR scheduler type           & Linear            \\ \hline
        Number of epochs            & 3                 \\ \hline
        \end{tabular}
    \end{minipage}
\end{table}

\newpage
\subsection{GO-MCTS Details} \;

In Algorithm \ref{alg:go-mcts-long}, we provide more details on GO-MCTS as it was implemented in this paper.

\begin{algorithm}[h]
\DontPrintSemicolon
\SetKwFunction{FMain}{GO-MCTS}
\SetKwFunction{FExpand}{Expand}
\SetKwFunction{FSample}{Sample}
\SetKwFunction{FArgMax}{Argmax}
\SetKwFunction{FRollout}{Rollout}
\SetKwFunction{FBackup}{Backup}
\SetKwFunction{FSelect}{Select}
\SetKwProg{Fn}{Function}{}{}
\begin{multicols}{2}
\Fn{\FMain{$h_{root}$, Tree, player, $N_{runs}$, $N_{steps}$, GM, $C$, $\mu$, threshold}}{
    \For{run $\leftarrow 1$ \KwTo $N_{runs}$}{
        $h \leftarrow h_{root}$\;
        \While{$h$ not Terminal}{
            \uIf{not ToMove($h$, player)}{
                $h$.Append(\FSample{$h$, GM})\;
            }
            \Else{
                \uIf{$h$ is in Tree}{
                    action = \FSelect{Tree, $h$, $C$}\;
                    $h$.Append(action)\;
                }
                \Else{
                    \FExpand{Tree, $h$, GM, threshold}\;
                    $h \leftarrow$ \FRollout{$h$, player, $N_{steps}$}\;
                    \textbf{break}\;
                }
            }
        }
        \FBackup{Tree, $h$, GM, $\mu$, player}\;
    }
    \KwRet \FArgMax{Tree[$h_{root}$]}\;
}
\vspace{5pt}
\Fn{\FSelect{Tree, $h$, $C$}}{
    totalVisits $\leftarrow$ 0\;
    \ForEach{$action$, ($val$, $visits$) in Tree[$h$].children}{
        totalVisits $\leftarrow$ totalVisits + $visits$\;
    }

    $maxScore \leftarrow -\infty$\;
    $bestAction \leftarrow$ NULL\;
    \ForEach{$action$, ($val$, $visits$) in Tree[$h$].children}{
        $uctScore$ $\leftarrow$ $\frac{val}{visits} + C \cdot \sqrt{\frac{\log(totalVisits)}{visits}}$\;
        \If{$uctScore > maxScore$}{
            $maxScore \leftarrow uctScore$\;
            $bestAction \leftarrow action$\;
        }
    }
    
    \KwRet{$bestAction$}\;
}

\columnbreak

\Fn{\FExpand{Tree, $h$, GM, threshold}}{
    node $\leftarrow$ NewNode()\;
    legalActions $\leftarrow$ LegalActions($h$)\;
    policy $\leftarrow$ GetLegalPolicy($h$, GM)\;
    \ForEach{action $\in$ legalActions}{
        \If{policy[action] $>$ threshold}{
            $h'$ = $h$.Append(action)\;
            $value$ $\leftarrow$ PredictValue($h'$, GM)\;
            node.children[action] $\leftarrow$ ($value$, 1)\;
        }
    }
    Tree[$h$] $\leftarrow$ node\;
}
\vspace{5pt}
\Fn{\FRollout{$h$, player, $N_{steps}$}}{
    \While{$N_{steps} > 0$ and $h$ not Terminal}{
        $h$.Append(\FSample{$h$, GM})\;
        \If{ToMove($h$, player)}{
            $N_{steps} \leftarrow N_{steps} - 1$\;
        }
    }
    \KwRet $h$\;
}
\vspace{5pt}
\Fn{\FBackup{Tree, $h$, GM, $\mu$, player}}{
    $value$ $\leftarrow$ OutcomeValue($h$, player) if $h$ is Terminal else PredictValue($h$, GM)\;
    $legal$ = IsLegal($h$)\;
    \While{$h \neq \emptyset$}{
        $a$ = $h$.PopBack()\;
        \If{$h$ in Tree}{
            node $\leftarrow$ Tree[$h$]\;
            (val, visits) $\leftarrow$ node.children[$a$]\;
            \uIf{legal}{
                node.children[$a$] $\leftarrow$ (val + $value$, visits + 1)\;
            }
            \Else{node.children[$a$] $\leftarrow$ (val - $\mu$, visits)\;
            }
        }
    }
}

\end{multicols}
\caption{Generative Observation Monte Carlo Tree Search (GO-MCTS) algorithm. $h_{root}$ represents the initial observation state, Tree is the search tree, player denotes the player for whom the decision is being made, $N_{runs}$ is the number of iterations, $N_{steos}$ is the maximum number of steps in the rollout, GM is the generative model used for simulation, $C$ is the exploration constant in UCT, $\mu$ is the penalty for illegal moves, and threshold is used to filter actions based on policy predictions. $\FSample$ simulates the next observation state, $\FArgMax$ selects the action with the highest value, $\FSelect$ chooses the action using UCT, $\FExpand$ adds new nodes to the tree, and $\FBackup$ updates node values based on rollout outcomes. Legality of history is assumed to be true for non-terminal histories.}
\label{alg:go-mcts-long}

\end{algorithm}

\subsection{Tournament Experimental Settings}

In our study, the state-based simulator was integrated into the Monte Carlo Tree Search (MCTS) framework, as described in Algorithm~\ref{alg:go-mcts}. This integration facilitated the identification of the active player and the legal actions available to the search player, specifically tailoring the decision-making process to the search player and excluding considerations for other players or environmental interactions.

The efficacy of the Generative Observation Monte Carlo Tree Search (GO-MCTS) player was evaluated through a series of tournaments in various card games, with distinct configurations for each tournament:

\begin{itemize}
    \item \textbf{Hearts Tournament Against Xinxin:} The GO-MCTS player was configured with 100 runs, a maximum of 2 rollouts, an exploration constant \(C = 0.4\), a policy action selection threshold of 0.05, and a penalty value \(\mu = 0.01\) for illegal moves. The comparison Argmax player had its selection threshold set to 0.05.
    
    \item \textbf{Skat Tournament Against Kermit:} For this tournament, the GO-MCTS player settings included 100 runs, up to 5 rollouts, an exploration constant \(C = 0.3\), with the policy action selection threshold and the penalty for illegal moves maintained at 0.05 and 0.01, respectively. The Argmax player's selection threshold was also set to 0.05.
    
    \item \textbf{Crew Tournament (Self-Play, No Baseline):} The GO-MCTS player was adjusted to 50 runs, 5 maximum rollouts, an exploration constant \(C = 0.1\), and a lower policy action selection threshold of 0.01, with the penalty for illegal moves \(\mu\) remaining at 0.01. The selection threshold for the Argmax player was set to 0.05.
\end{itemize}

\twocolumn

\bibliography{main.bib}

\begin{thebibliography}{42}
\providecommand{\natexlab}[1]{#1}

\bibitem[{Antonoglou et~al.(2021)Antonoglou, Schrittwieser, Ozair, Hubert, and
  Silver}]{antonoglou2021planning}
Antonoglou, I.; Schrittwieser, J.; Ozair, S.; Hubert, T.~K.; and Silver, D.
  2021.
\newblock Planning in Stochastic Environments with a Learned Model.
\newblock In \emph{International Conference on Learning Representations}.

\bibitem[{{\AA}str{\"o}m(1965)}]{aastrom1965optimal}
{\AA}str{\"o}m, K.~J. 1965.
\newblock Optimal Control of Markov Processes with Incomplete State Information
  I.
\newblock \emph{Journal of mathematical analysis and applications}, 10:
  174--205.

\bibitem[{Azizzadenesheli et~al.(2018)Azizzadenesheli, Yang, Liu, Lipton, and
  Anandkumar}]{azizzadenesheli2018surprising}
Azizzadenesheli, K.; Yang, B.; Liu, W.; Lipton, Z.~C.; and Anandkumar, A. 2018.
\newblock Surprising Negative Results for Generative Adversarial Tree Search.
\newblock \emph{arXiv preprint arXiv:1806.05780}.

\bibitem[{Bansal et~al.(2017)Bansal, Pachocki, Sidor, Sutskever, and
  Mordatch}]{bansal2017emergent}
Bansal, T.; Pachocki, J.; Sidor, S.; Sutskever, I.; and Mordatch, I. 2017.
\newblock Emergent Complexity via Multi-Agent Competition.
\newblock \emph{arXiv preprint arXiv:1710.03748}.

\bibitem[{Brown and Sandholm(2018)}]{brown2018superhuman}
Brown, N.; and Sandholm, T. 2018.
\newblock Superhuman AI for Heads-up No-limit Poker: Libratus Beats Top
  Professionals.
\newblock \emph{Science}, 359(6374): 418--424.

\bibitem[{Buro et~al.(2009)Buro, Long, Furtak, and
  Sturtevant}]{buro2009improving}
Buro, M.; Long, J.~R.; Furtak, T.; and Sturtevant, N. 2009.
\newblock Improving State Evaluation, Inference, and Search in Trick-Based Card
  Games.
\newblock In \emph{Twenty-First International Joint Conference on Artificial
  Intelligence}.

\bibitem[{Campbell, Hoane~Jr, and Hsu(2002)}]{campbell2002deep}
Campbell, M.; Hoane~Jr, A.~J.; and Hsu, F.-h. 2002.
\newblock Deep Blue.
\newblock \emph{Artificial intelligence}, 134(1-2): 57--83.

\bibitem[{Chen et~al.(2021)Chen, Lu, Rajeswaran, Lee, Grover, Laskin, Abbeel,
  Srinivas, and Mordatch}]{chen2021decision}
Chen, L.; Lu, K.; Rajeswaran, A.; Lee, K.; Grover, A.; Laskin, M.; Abbeel, P.;
  Srinivas, A.; and Mordatch, I. 2021.
\newblock Decision transformer: Reinforcement Learning via Sequence Modeling.
\newblock \emph{Advances in neural information processing systems}, 34:
  15084--15097.

\bibitem[{Cowling, Powley, and Whitehouse(2012)}]{cowling2012information}
Cowling, P.~I.; Powley, E.~J.; and Whitehouse, D. 2012.
\newblock Information Set Monte Carlo Tree Search.
\newblock \emph{IEEE Transactions on Computational Intelligence and AI in
  Games}, 4(2): 120--143.

\bibitem[{Deisenroth et~al.(2013)Deisenroth, Neumann, Peters
  et~al.}]{deisenroth2013survey}
Deisenroth, M.~P.; Neumann, G.; Peters, J.; et~al. 2013.
\newblock A Survey on Policy Search for Robotics.
\newblock \emph{Foundations and Trends{\textregistered} in Robotics}, 2(1--2):
  1--142.

\bibitem[{Frank and Basin(1998)}]{frank1998search}
Frank, I.; and Basin, D. 1998.
\newblock Search in Games with Incomplete Information: A Case Study using
  Bridge Card Play.
\newblock \emph{Artificial Intelligence}, 100(1-2): 87--123.

\bibitem[{Gerhardt(2004)}]{gerhardt_2004}
Gerhardt, G. 2004.

\bibitem[{Ginsberg(2001)}]{ginsberg2001gib}
Ginsberg, M.~L. 2001.
\newblock GIB: Imperfect Information in a Computationally Challenging Game.
\newblock \emph{Journal of Artificial Intelligence Research}, 14: 303--358.

\bibitem[{Heinrich and Silver(2016)}]{heinrich2016deep}
Heinrich, J.; and Silver, D. 2016.
\newblock Deep Reinforcement Learning from Self-Play in Imperfect-Information
  Games.
\newblock \emph{arXiv preprint arXiv:1603.01121}.

\bibitem[{Janner, Li, and Levine(2021)}]{janner2021offline}
Janner, M.; Li, Q.; and Levine, S. 2021.
\newblock Offline reinforcement Learning as one Big Sequence Modeling Problem.
\newblock \emph{Advances in neural information processing systems}, 34:
  1273--1286.

\bibitem[{Kaelbling, Littman, and Cassandra(1998)}]{kaelbling1998planning}
Kaelbling, L.~P.; Littman, M.~L.; and Cassandra, A.~R. 1998.
\newblock Planning and Acting in Partially Observable Stochastic Domains.
\newblock \emph{Artificial intelligence}, 101(1-2): 99--134.

\bibitem[{Kearns, Mansour, and Ng(2002)}]{kearns2002sparse}
Kearns, M.; Mansour, Y.; and Ng, A.~Y. 2002.
\newblock A Sparse Sampling Algorithm for Near-Optimal Planning in Large Markov
  Decision Processes.
\newblock \emph{Machine learning}, 49(2): 193--208.

\bibitem[{Khan and Sun(2002)}]{khan2002non}
Khan, M.~A.; and Sun, Y. 2002.
\newblock Non-Cooperative Games with Many Players.
\newblock \emph{Handbook of game theory with economic applications}, 3:
  1761--1808.

\bibitem[{Kocsis and Szepesv{\'a}ri(2006)}]{kocsis2006bandit}
Kocsis, L.; and Szepesv{\'a}ri, C. 2006.
\newblock Bandit Based Monte-Carlo Planning.
\newblock In \emph{European conference on machine learning}, 282--293.
  Springer.

\bibitem[{Lanctot et~al.(2019)Lanctot, Lockhart, Lespiau, Zambaldi, Upadhyay,
  P\'{e}rolat, Srinivasan, Timbers, Tuyls, Omidshafiei, Hennes, Morrill,
  Muller, Ewalds, Faulkner, Kram\'{a}r, Vylder, Saeta, Bradbury, Ding,
  Borgeaud, Lai, Schrittwieser, Anthony, Hughes, Danihelka, and
  Ryan-Davis}]{LanctotEtAl2019OpenSpiel}
Lanctot, M.; Lockhart, E.; Lespiau, J.-B.; Zambaldi, V.; Upadhyay, S.;
  P\'{e}rolat, J.; Srinivasan, S.; Timbers, F.; Tuyls, K.; Omidshafiei, S.;
  Hennes, D.; Morrill, D.; Muller, P.; Ewalds, T.; Faulkner, R.; Kram\'{a}r,
  J.; Vylder, B.~D.; Saeta, B.; Bradbury, J.; Ding, D.; Borgeaud, S.; Lai, M.;
  Schrittwieser, J.; Anthony, T.; Hughes, E.; Danihelka, I.; and Ryan-Davis, J.
  2019.
\newblock {OpenSpiel}: A Framework for Reinforcement Learning in Games.
\newblock \emph{CoRR}, abs/1908.09453.

\bibitem[{Levy(1989)}]{levy1989million}
Levy, D. 1989.
\newblock The Million Pound Bridge Program, Heuristic Programming in Artificial
  Intelligence: The First Computer Olympiad.

\bibitem[{Li et~al.(2022)Li, Hopkins, Bau, Vi{\'e}gas, Pfister, and
  Wattenberg}]{li2022emergent}
Li, K.; Hopkins, A.~K.; Bau, D.; Vi{\'e}gas, F.; Pfister, H.; and Wattenberg,
  M. 2022.
\newblock Emergent World Representations: Exploring a Sequence Model Trained on
  a Synthetic Task.
\newblock \emph{arXiv preprint arXiv:2210.13382}.

\bibitem[{Long et~al.(2010)Long, Sturtevant, Buro, and
  Furtak}]{long2010understanding}
Long, J.~R.; Sturtevant, N.~R.; Buro, M.; and Furtak, T. 2010.
\newblock Understanding the Success of Perfect Information Monte Carlo Sampling
  in Game Tree Search.
\newblock In \emph{Twenty-Fourth AAAI Conference on Artificial Intelligence}.

\bibitem[{Loshchilov and Hutter(2017)}]{loshchilov2017decoupled}
Loshchilov, I.; and Hutter, F. 2017.
\newblock Decoupled Weight Decay Regularization.
\newblock \emph{arXiv preprint arXiv:1711.05101}.

\bibitem[{Morav{\v{c}}{\'\i}k et~al.(2017)Morav{\v{c}}{\'\i}k, Schmid, Burch,
  Lis{\`y}, Morrill, Bard, Davis, Waugh, Johanson, and
  Bowling}]{moravvcik2017deepstack}
Morav{\v{c}}{\'\i}k, M.; Schmid, M.; Burch, N.; Lis{\`y}, V.; Morrill, D.;
  Bard, N.; Davis, T.; Waugh, K.; Johanson, M.; and Bowling, M. 2017.
\newblock DeepStack: Expert-Level Artificial Intelligence in Heads-Up No-Limit
  Poker.
\newblock \emph{Science}, 356(6337): 508--513.

\bibitem[{Ozair et~al.(2021)Ozair, Li, Razavi, Antonoglou, Van Den~Oord, and
  Vinyals}]{ozair2021vector}
Ozair, S.; Li, Y.; Razavi, A.; Antonoglou, I.; Van Den~Oord, A.; and Vinyals,
  O. 2021.
\newblock Vector Quantized Models for Planning.
\newblock In \emph{international conference on machine learning}, 8302--8313.
  PMLR.

\bibitem[{Paszke et~al.(2019)Paszke, Gross, Massa, Lerer, Bradbury, Chanan,
  Killeen, Lin, Gimelshein, Antiga, Desmaison, Kopf, Yang, DeVito, Raison,
  Tejani, Chilamkurthy, Steiner, Fang, Bai, and Chintala}]{pytiorch}
Paszke, A.; Gross, S.; Massa, F.; Lerer, A.; Bradbury, J.; Chanan, G.; Killeen,
  T.; Lin, Z.; Gimelshein, N.; Antiga, L.; Desmaison, A.; Kopf, A.; Yang, E.;
  DeVito, Z.; Raison, M.; Tejani, A.; Chilamkurthy, S.; Steiner, B.; Fang, L.;
  Bai, J.; and Chintala, S. 2019.
\newblock PyTorch: An Imperative Style, High-Performance Deep Learning Library.
\newblock In \emph{Advances in Neural Information Processing Systems 32},
  8024--8035. Curran Associates, Inc.

\bibitem[{Radford et~al.(2019)Radford, Wu, Child, Luan, Amodei, Sutskever
  et~al.}]{radford2019language}
Radford, A.; Wu, J.; Child, R.; Luan, D.; Amodei, D.; Sutskever, I.; et~al.
  2019.
\newblock Language Models are Unsupervised Multitask Learners.
\newblock \emph{OpenAI blog}, 1(8): 9.

\bibitem[{Schaeffer et~al.(2007)Schaeffer, Burch, Bjornsson, Kishimoto, Muller,
  Lake, Lu, and Sutphen}]{schaeffer2007checkers}
Schaeffer, J.; Burch, N.; Bjornsson, Y.; Kishimoto, A.; Muller, M.; Lake, R.;
  Lu, P.; and Sutphen, S. 2007.
\newblock Checkers is Solved.
\newblock \emph{science}, 317(5844): 1518--1522.

\bibitem[{Schrittwieser et~al.(2020)Schrittwieser, Antonoglou, Hubert,
  Simonyan, Sifre, Schmitt, Guez, Lockhart, Hassabis, Graepel
  et~al.}]{schrittwieser2020mastering}
Schrittwieser, J.; Antonoglou, I.; Hubert, T.; Simonyan, K.; Sifre, L.;
  Schmitt, S.; Guez, A.; Lockhart, E.; Hassabis, D.; Graepel, T.; et~al. 2020.
\newblock Mastering Atari, Go, Chess and Shogi by Planning with a Learned
  Model.
\newblock \emph{Nature}, 588(7839): 604--609.

\bibitem[{Seify and Buro(2020)}]{seify2020single}
Seify, A.; and Buro, M. 2020.
\newblock Single-Agent Optimization Through Policy Iteration using Monte-Carlo
  Tree Search.
\newblock \emph{arXiv preprint arXiv:2005.11335}.

\bibitem[{Silver et~al.(2016)Silver, Huang, Maddison, Guez, Sifre, Van
  Den~Driessche, Schrittwieser, Antonoglou, Panneershelvam, Lanctot
  et~al.}]{silver2016mastering}
Silver, D.; Huang, A.; Maddison, C.~J.; Guez, A.; Sifre, L.; Van Den~Driessche,
  G.; Schrittwieser, J.; Antonoglou, I.; Panneershelvam, V.; Lanctot, M.;
  et~al. 2016.
\newblock Mastering the Game of Go with Deep Neural Networks and Tree Search.
\newblock \emph{nature}, 529(7587): 484--489.

\bibitem[{Silver and Veness(2010)}]{silver2010monte}
Silver, D.; and Veness, J. 2010.
\newblock Monte-Carlo Planning in Large POMDPs.
\newblock \emph{Advances in neural information processing systems}, 23.

\bibitem[{Sing(2020)}]{sing2000crew}
Sing, T. 2020.
\newblock The Crew: The Quest for Planet Nine.

\bibitem[{Solinas et~al.(2023)Solinas, Rebstock, Sturtevant, and
  Buro}]{solinas2023history}
Solinas, C.; Rebstock, D.; Sturtevant, N.~R.; and Buro, M. 2023.
\newblock History Filtering in Imperfect Information Games: Algorithms and
  Complexity.
\newblock \emph{arXiv preprint arXiv:2311.14651}.

\bibitem[{Sturtevant(2004)}]{sturtevant2004current}
Sturtevant, N. 2004.
\newblock Current challenges in multi-player game search.
\newblock In \emph{International Conference on Computers and Games}, 285--300.
  Springer.

\bibitem[{Sturtevant(2008)}]{sturtevant2008analysis}
Sturtevant, N. 2008.
\newblock An analysis of UCT in multi-player games.
\newblock \emph{ICGA Journal}, 31(4): 195--208.

\bibitem[{Sturtevant, Zinkevich, and Bowling(2006)}]{sturtevant2006prob}
Sturtevant, N.; Zinkevich, M.; and Bowling, M. 2006.
\newblock Prob-max\^{} n: Playing N-player Games with Opponent Models.
\newblock In \emph{AAAI}, volume~6, 1057--1063.

\bibitem[{Sturtevant and White(2007)}]{sturtevant2007feature}
Sturtevant, N.~R.; and White, A.~M. 2007.
\newblock Feature Construction for Reinforcement Learning in Hearts.
\newblock In \emph{Computers and Games: 5th International Conference, CG 2006,
  Turin, Italy, May 29-31, 2006. Revised Papers 5}, 122--134. Springer.

\bibitem[{Sutton(1990)}]{sutton1990integrated}
Sutton, R.~S. 1990.
\newblock Integrated Architectures for Learning, Planning, and Reacting Based
  on Approximating Dynamic Programming.
\newblock In \emph{Machine learning proceedings 1990}, 216--224. Elsevier.

\bibitem[{Wolf et~al.(2019)Wolf, Debut, Sanh, Chaumond, Delangue, Moi, Cistac,
  Rault, Louf, Funtowicz et~al.}]{wolf2019huggingface}
Wolf, T.; Debut, L.; Sanh, V.; Chaumond, J.; Delangue, C.; Moi, A.; Cistac, P.;
  Rault, T.; Louf, R.; Funtowicz, M.; et~al. 2019.
\newblock Huggingface's Transformers: State-of-the-Art Natural Language
  Processing.
\newblock \emph{arXiv preprint arXiv:1910.03771}.

\bibitem[{Yaari et~al.(2022)Yaari, Rokach, Puzis, and
  Katz}]{yaari2022mctransformer}
Yaari, G.; Rokach, L.; Puzis, R.; and Katz, G. 2022.
\newblock MCTransformer: Combining Transformers and Monte-Carlo Tree Search for
  Offline Reinforcement Learning.

\end{thebibliography}

\end{document}